\documentclass[journal]{IEEEtran}
\usepackage[hidelinks]{hyperref}

\ifCLASSINFOpdf
\else
\fi
\usepackage{amsmath}
\usepackage{mathtools}
\usepackage{algorithm}
\usepackage{algpseudocode}
\algnewcommand{\LineComment}[1]{\State \textcolor{teal}{\(\triangleright\) #1}}

\algdef{SE}[SUBALG]{Indent}{EndIndent}{}{\algorithmicend\ }%
\algtext*{Indent}
\algtext*{EndIndent}
\usepackage{bm}

\usepackage{xcolor}
\usepackage{graphicx}
\DeclareMathOperator*{\argmin}{arg\,min} 
\usepackage{subfig}
\usepackage{boldline}
\usepackage{amssymb}
\usepackage{booktabs}
\usepackage{multirow}

\usepackage{pifont}

\begin{document}
%

\title{Resource-Efficient Federated Multimodal Learning via Layer-wise and Progressive Training
}

\author{Ye Lin Tun, Chu Myaet Thwal, Minh N. H. Nguyen, and Choong Seon Hong,~\IEEEmembership{Fellow,~IEEE}
\thanks{Ye Lin Tun, Chu Myaet Thwal, and Choong Seon Hong are with the Department of Computer Science and Engineering, Kyung Hee University, Yongin-si, Gyeonggi-do 17104, Republic of Korea (e-mail: \{yelintun, chumyaet, cshong\}@khu.ac.kr). }
\thanks{Minh N. H. Nguyen is with the Vietnam - Korea University of Information and Communication Technology, Danang, Vietnam (e-mail: nhnminh@vku.udn.vn). }
}

\maketitle

\begin{abstract}

Combining different data modalities enables deep neural networks to tackle complex tasks more effectively, making multimodal learning increasingly popular. 
To harness multimodal data closer to end users, it is essential to integrate multimodal learning with privacy-preserving approaches like federated learning (FL).
However, compared to conventional unimodal learning, multimodal setting requires dedicated encoders for each modality, resulting in larger and more complex models. Training these models requires significant resources, presenting a substantial challenge for FL clients operating with limited computation and communication resources.
To address these challenges, we introduce LW-FedMML, a layer-wise federated multimodal learning approach which decomposes the training process into multiple stages.
Each stage focuses on training only a portion of the model, thereby significantly  reducing the memory and computational requirements.
Moreover, FL clients only need to exchange the trained model portion with the central server, lowering the resulting communication cost.
We conduct extensive experiments across various FL and multimodal learning settings to validate the effectiveness of our proposed method. 
The results demonstrate that LW-FedMML can compete with conventional end-to-end federated multimodal learning \mbox{(FedMML)} while significantly reducing the resource burden on FL clients. 
Specifically, \mbox{LW-FedMML} reduces memory usage by up to $2.7\times$, computational operations (FLOPs) by $2.4\times$, and total communication cost by $2.3\times$.
We also explore a progressive training approach called Prog-FedMML.
While it offers lesser resource efficiency than LW-FedMML, \mbox{Prog-FedMML} has the potential to surpass the performance of end-to-end FedMML, making it a viable option for scenarios with fewer resource constraints.

\end{abstract}


%
\IEEEpeerreviewmaketitle

\section{Introduction}

\IEEEPARstart{D}{eep} neural networks have become an integral part of many practical applications in our daily lives. 
While these models have traditionally been trained on a single modality, recent advancements in the field are driving a shift towards multimodal learning \cite{10123038, bayoudh2022survey, pmlr-v139-radford21a}. 
This transition is motivated by the realization that leveraging multiple types of data simultaneously can significantly enhance the performance of the models \cite{8269806}.
Just as animal intelligence relies on multiple senses to perceive their environment, combining different data modalities allows models to tackle complex tasks more effectively. 
Integrating diverse data sources, such as vision and audio, ensures that more information is available to the model, thereby improving its perception and decision-making process \cite{ngiam2011multimodal}.

Multimodal learning based on instance discrimination approaches aligns representations of different modalities within the same data sample, thereby learning transferable representations \cite{pmlr-v139-radford21a, elizalde2023clap}. 
On the other hand, supervised multimodal learning aims to fuse information from various modalities to enhance the predictive performance of the model \cite{8103116, ofli2013berkeley, hu2020cross}.
As mobile and IoT devices become increasingly powerful, they are integrated with a broader range of sensors \cite{radu2018multimodal}.
For instance, smartphones are equipped with cameras, gyroscopes, accelerometers, and more, while smart home IoT devices can include an array of modular sensors for various functionalities.
These edge devices capture and store real-world multimodal data, which is valuable in training AI models for practical applications.
However, conventional multimodal learning methods require collecting this data into a central repository for training, raising data privacy concerns.

To harness the distributed data while also considering privacy issues, numerous studies have explored decentralized training approaches like federated learning (FL) \cite{mcmahan2017communication, konevcny2016federated, kairouz2021advances, yang2019federated}.
In an FL system, the central server collects model parameters trained on the private data of edge devices rather than the data itself \cite{mcmahan2017communication}. 
Such a mechanism effectively guarantees the confidentiality of data held by different parties involved in the training process.
However, this also means that the FL training process relies heavily on the computation and communication capabilities of FL clients, as they are responsible for model training \cite{yu2021toward}.

While a single encoder is sufficient to process the data in unimodal tasks, multimodal tasks often require dedicated encoders tailored for each modality. 
Consequently, multimodal learning demands significant computational and memory resources to accommodate these modality-specific encoders.
This presents a drawback for FL clients, which are often edge devices with limited resources, making it difficult to train multimodal models \cite{che2023multimodal}.
The problem becomes even more challenging when dealing with high-dimensional data, such as video or audio.

The need for modality-specific encoders makes multimodal models larger and more complex than their unimodal counterparts, leading to increased transmission costs.
This creates communication bottlenecks in the FL process, as FL clients may struggle to exchange these large multimodal models with the central server during training. 
Due to diverse hardware capabilities and resource constraints, many FL clients may not meet the resource requirements for conventional end-to-end multimodal training. As a result, their valuable data is excluded from the training process. 
Therefore, exploring resource-efficient approaches is crucial for maximizing the effectiveness of multimodal learning within the FL context.

During training, reducing the batch size may appear to be a straightforward solution to lower the computational demands on resource-constrained devices. 
However, in instance discrimination-based multimodal learning approaches (e.g., CLIP~\cite{pmlr-v139-radford21a}), a small batch size may compromise the effectiveness of the contrastive loss, degrading the quality of learned representations \cite{chen2020simple, li2023mocosfl}.
These methods benefit from a larger batch size to increase the number of negative samples, which is essential for the contrastive loss to perform well. 
Furthermore, simply reducing the batch size only addresses the computational concerns but fails to alleviate the communication burden on FL clients, as they still need to exchange the same model size with the server.

To address the resource constraints in an FL environment for multimodal learning, our work introduces the layer-wise training approach.
This approach decomposes the training process into multiple stages, with each stage dedicated to training a single layer or a portion of the model.
By training different layers in separate stages, we can effectively reduce the computational and memory requirements on FL clients.
Furthermore, at each stage, only the layer active for training needs to be exchanged with the server, thereby reducing the communication burden on FL clients.

Our contributions are summarized as follows: (i) We introduce LW-FedMML, a layer-wise federated multimodal learning strategy designed to address the resource limitations of FL clients engaged in multimodal tasks.
LW-FedMML can match the performance of conventional end-to-end training while significantly reducing resource requirements.
(ii) We also explore a progressive training approach, referred to as Prog-FedMML.
Although Prog-FedMML does not offer the same level of resource efficiency as LW-FedMML, it has the potential to outperform conventional end-to-end training while maintaining similar memory usage and lowering computational and communication costs.
(iii) To validate the effectiveness of our proposed methods, we conduct extensive experiments across various FL scenarios and multimodal learning setups.

\section{Preliminary}

\subsection{Federated Learning}

\begin{figure}[tb]
    \centering
    \includegraphics[width=0.7\linewidth]{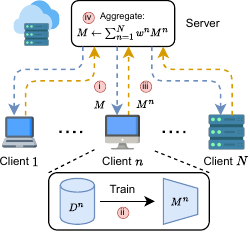}
    \caption{Overview of the federated learning process. (i) The global model $M$ is distributed to clients. (ii) Each client $n$ performs training on the local dataset $D^n$. (iii) The local model $M^n$ is sent back to the server. (iv) The server performs the aggregation.}
    \label{fig:fl}
\end{figure}

A federated learning (FL) process typically involves a central server and a set of FL client devices \cite{mcmahan2017communication, li2020federated, zhao2018federated}. 
These devices are often found close to end-users, capturing and holding valuable data for model training.
Meanwhile, the central server oversees and monitors the FL procedure. 
Given $N$ clients, the goal of FL is to train a global model $M$ by solving the following optimization problem:
\begin{equation}
    M^* = \argmin_{M} \sum_{n=1}^N w^n \mathcal{L} (M, D^n),
    \label{eqn:fl_obj}
\end{equation}
where $\mathcal{L}$ denotes the loss function and $D^n$ represents the local dataset of the $n$-th client.
Here, the weight $w^n$ assigned to client $n$ is typically calculated as:
\begin{equation}
w^n=\frac{|D^n|}{|D|} \ \text{,} \ \text{where} \ {D=\bigcup_{n=1}^{N} D^n},
\label{eqn:weight_i}
\end{equation}
with $|D^n|$ and $|D|$ representing the number of data samples in $D^n$ and $D$, respectively.
As illustrated in Fig.~\ref{fig:fl}, four key steps comprise a standard FL communication round. 
(i) The central server distributes the global model $M$ to each participating client $n \in [1,N]$. 
(ii) Each client $n$ trains the received model on its local dataset $D^n$ to produce the local model $M^n$. 
(iii) The trained local models $[M^1, \dots, M^N]$ are sent back to the server. 
(iv) The server aggregates the received local models and updates the global model. 
The FedAvg~\cite{mcmahan2017communication} algorithm is commonly used for the aggregation process, which can be described as  $M \leftarrow \sum^N_{n=1} w^n M^n$. 
These four steps are iteratively executed until the model converges or a predefined number of FL communication rounds $R$ is reached.

\subsection{Multimodal Learning}

\begin{figure}[tb]
    \centering
    \subfloat[Instance discrimination-based multimodal learning.]{\includegraphics[width=0.85\linewidth]{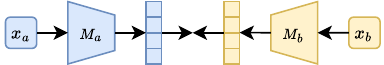} \label{subfig:mml_clip}} \\
	\subfloat[Supervised multimodal learning.]{\includegraphics[width=0.85\linewidth]{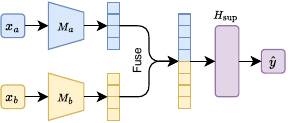} \label{subfig:mml_sup}}
	\caption{Common multimodal learning approaches given modalities $a$ and $b$. $M$ denotes the model, and $x$ denotes the input sample. (a) The instance discrimination-based approach focuses on aligning the representations of diverse modalities within a given data sample. Negative samples are omitted in the figure for clarity. (b) The supervised approach aims to fuse information from various modalities, enabling the model to make more informed predictions. Here, $H_\text{sup}$ can be any task-specific prediction head.}
	\label{fig:mml}
\end{figure}

In many complex tasks, such as healthcare, autonomous driving, and weather forecasting, a single modality may not provide sufficient context or detail for accurate decision-making \cite{muhammad2021comprehensive, caesar2020nuscenes, bai2022multimodal}. 
Multimodal learning is essential in these scenarios, as it leverages the complementary information of different data modalities, providing a more comprehensive and detailed understanding of the task at hand.
Many studies on multimodal learning center around aligning the representations of diverse modalities within a given data sample, as illustrated in Fig.~\ref{subfig:mml_clip}. 
These approaches often employ contrastive loss, as seen in CLIP~\cite{pmlr-v139-radford21a}, to draw representations of different modalities within the same sample  closer together, while simultaneously pushing those from different samples further apart.
This allows encoders of different modalities to learn transferable representations applicable to a wide range of downstream tasks.
These pretrained encoders can then be deployed either individually or in pairs, depending on whether the downstream tasks involve unimodality or multimodality processing.
On the other hand, supervised multimodal learning, depicted in Fig.~\ref{subfig:mml_sup}, seeks to fuse information from various modalities, enabling the model to make more accurate and informed predictions \cite{8103116, ofli2013berkeley, hu2020cross}.
A straightforward way to fuse representations of different modalities is to concatenate them, thereby creating a unified feature representation that contains the essence of each modality.
However, supervised approaches rely on labeled data for training, which can be costly and labor-intensive to obtain, often serving as a limiting factor in many cases. 

\section{Related Work}
\label{sec:related_work}

Many studies have shown increasing interest in federated multimodal learning due to its potential to be more powerful than conventional unimodal approaches \cite{che2023multimodal}.
In \cite{xiong2022unified}, the authors propose a co-attention mechanism to fuse information from different modalities within clients.
Hierarchical gradient blending is introduced in \cite{chen2022towards} to address the challenge of varying generalization rates among modality sub-networks and local models. 
The authors in \cite{liu2020federated} aim to learn fine-grained image representations from vision and language modalities in the FL setting.
In the medical field, \cite{agbley2021multimodal} demonstrates the application of federated multimodal learning to detect melanoma from skin lesion images and clinical data.
Similarly, \cite{bernecker2022fednorm} utilizes CT and MRI images for liver segmentation.
These studies highlight the versatility and potential of federated multimodal learning to leverage diverse data sources in various domains.

The layer-wise training approach was originally employed to train deep belief networks and restricted Boltzmann machines \cite{10.1162/neco.2006.18.7.1527,10.5555/2976456.2976476}, to avoid training issues such as vanishing gradients.
This approach can significantly reduce resource requirements in the context of federated multimodal learning, where the integration of multiple modalities imposes even greater computational and communication demands compared to unimodal FL.
The authors in \cite{wang2022progfed} propose a progressive training approach that incrementally trains the model by increasing its depth. Similarly, \cite{tun2024lw} and \cite{pengfei2023towards} explore layer-wise training approaches for federated self-supervised learning.
In contrast, these aforementioned studies primarily consider unimodal settings. 
Our work extends layer-wise and progressive training approaches to the multimodal context in FL, addressing the unique challenges and opportunities presented by multimodal data in FL environments.

\section{Proposed Method}
\label{sec:method}

\begin{figure}[tb]
    \centering
    \includegraphics[width=0.9\linewidth]{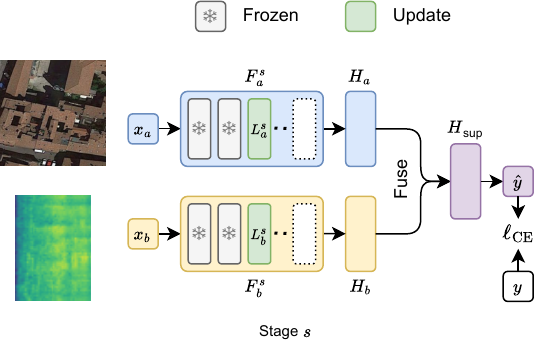}
    \caption{Overview of LW-FedMML for supervised setting at stage $s$.}
    \label{fig:sup_lwmml}
\end{figure}

\subsection{LW-FedMML}

The main idea of layer-wise training is to break down the training process of a model into multiple stages, denoted as ${s\in[1,S]}$.
Initially, a given encoder $F^0$ starts off without any layers.
At the beginning of each stage $s$, a new layer $L^s$ is sequentially attached to the encoder $F^s$, thereby increasing its depth. 
During stage $s$, only the corresponding layer $L^s$ is trained, while the other layers $[L^1, \dots , L^{(s-1)}]$ from previous stages are frozen without receiving updates.
Each stage $s$ lasts for a specified number of epochs, or communication rounds in the context of FL.

\begin{figure*}[tb]
    \centering
    \includegraphics[width=\linewidth]{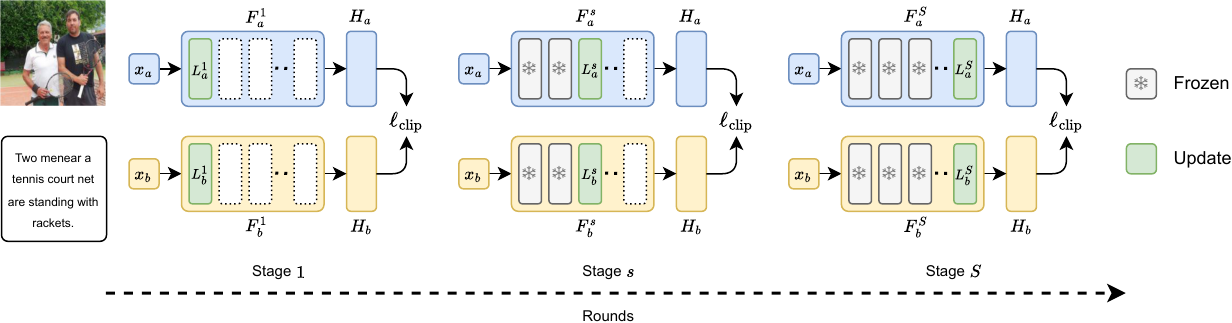}
    \caption{Overview of LW-FedMML for the instance discrimination-based setting.
    The training process is divided into multiple stages $s \in [1,S]$. 
    In stage $s$, the active layers $L_a^s$ and $L_b^s$ are depicted in green within encoders $F_a^s$ and $F_b^s$, where $a$ and $b$ represent different modalities. Prior layers within the encoders are frozen, as indicated in gray.}
    \label{fig:lwmml}
\end{figure*}

\begin{algorithm}[tb]
\caption{LW-FedMML (\emph{Server-side})}
\label{alg:server}
\textbf{Input:} encoder $F_a^0$ and projection head $H_a$ for modality $a$, encoder $F_b^0$ and projection head $H_b$ for modality $b$, number of stages $S$, number of rounds $R^s$ where  $s \in [1, S]$ \\
\textbf{Output:} $F_a^S$, $F_b^S$, $H_a$, $H_b$

\begin{algorithmic}[1]    
    \State \textbf{Server executes:}

    \For{stage $s=1,2,\dots, S$}
        \State Initialize new layers: $L_a^s$ and $L_b^s$
        \vspace{0.2em}
        \For{round $r=1,2,\dots, R^s$}
            \For{client $n=1,2,\dots,N$ in parallel}
                \vspace{0.2em}
                \State $L_a^{s,n}, L_b^{s,n}, H_a^n, H_b^n \leftarrow \text{Train}(L_a^s, L_b^s, H_a, H_b)$
                \vspace{0.2em}
                \State $w^n = \frac{|D^n|}{|D|}$, where $D=\bigcup_{n=1}^{N} D^n$

            \EndFor

            \vspace{0.2em}
            \State $L_a^s \leftarrow \sum_{n=1}^N w^n L_a^{s,n}, \quad H_a \leftarrow \sum_{n=1}^N w^n H_a^n$
            \vspace{0.2em}
            \State $L_b^s \leftarrow \sum_{n=1}^N w^n L_b^{s,n}, \quad H_b \leftarrow \sum_{n=1}^N w^n H_b^n$
            
        \EndFor
        \State Distribute $L_a^s$ and $L_b^s$ to clients
        \vspace{0.2em}
        \State $F_a^s \leftarrow$ Attach $L_a^s$ to $F_a^{(s-1)}$
        \vspace{0.2em}
        \State $F_b^s \leftarrow$ Attach $L_b^s$ to $F_b^{(s-1)}$
        \vspace{0.2em}
    \EndFor
    \State return $F_a^S$, $F_b^S$, $H_a$, $H_b$
    
\end{algorithmic}
\end{algorithm}

\begin{algorithm}[tb]
\caption{LW-FedMML (\emph{Client-side})}
\label{alg:client}
\textbf{Input:} local dataset $D^n$, layers $[L_a^1, \dots, L_a^{(s-1)}]$ for modality~$a$, layers $[L_b^1, \dots, L_b^{(s-1)}]$ for modality $b$, number of local epochs~$E$  \\
\textbf{Output:} $L_a^{s,n}, L_b^{s,n}, H_a^n, H_b^n$

\begin{algorithmic}[1]
    \State \textbf{Client executes:} $\text{Train}(L_a^s, L_b^s, H_a, H_b)$:
    \vspace{0.2em}
    \State $L_a^{s,n} \leftarrow L_a^s, \quad H_a^n \leftarrow H_a$
    \vspace{0.2em}
    \State $L_b^{s,n} \leftarrow L_b^s, \quad H_b^n \leftarrow H_b$
    \vspace{0.2em}
    \State $F_a^{s,n} \leftarrow$ Freeze $[L_a^1, \dots, L_a^{(s-1)}]$ and attach $L_a^{s,n}$
    \vspace{0.2em}
    \State $F_b^{s,n} \leftarrow$ Freeze $[L_b^1, \dots, L_b^{(s-1)}]$ and attach $L_b^{s,n}$
    \vspace{0.2em}
    \For{epoch $e=1,2,\dots,E$}
        \For{each batch $(x_a, x_b) \in D^n$}
            \vspace{0.2em}
            \State $z_a \leftarrow H_a^n(F_a^{s,n}(x_a))$
            \vspace{0.2em}
            \State $z_b \leftarrow H_b^n(F_b^{s,n}(x_b))$
            \vspace{0.2em}
            \State $\mathcal{L} \leftarrow \ell_{\text{clip}}(z_a, z_b)$
            \vspace{0.2em}
            \State $L_a^{s,n}, L_b^{s,n}, H_a^n, H_b^n \leftarrow$ Update with $\nabla\mathcal{L}$
            \vspace{0.2em}
        \EndFor
    \EndFor
    \State Upload $L_a^{s,n}, L_b^{s,n}, H_a^n, H_b^n$
\end{algorithmic}
\end{algorithm}

Fig.~\ref{fig:lwmml} illustrates the training process for LW-FedMML.
Specifically, given different modalities $a$ and $b$, layers $L^s_a$ and $L^s_b$ are attached to encoders $F_a^s$ and $F_b^s$ at the beginning of each stage $s$.
In instance discrimination-based \mbox{LW-FedMML}, we follow the CLIP~\cite{pmlr-v139-radford21a} approach, which computes the \mbox{InfoNCE}~\cite{van2018representation} loss for each modality.
This approach allows effective alignment of representations from different data modalities with simplicity.
Given a batch of paired samples $x_a$ and $x_b$, the projections are extracted as $z_a = H_a(F_a^s(x_a))$ and $z_b = H_b(F_b^s(x_b))$, where $H_a$ and $H_b$ denote modality-specific projection heads.
When both $z_a$ and $z_b$ are normalized, the loss for modality $a$ can be calculated as follows: 
\begin{equation}
    \label{eq:infonce}
    \ell(z_a, z_b) = -\frac{1}{B} \sum_{i=1}^B \log \frac{\exp(z_a^i \cdot z_b^i /\tau)}{\sum_{j=1}^B \exp(z_a^i \cdot z_b^j /\tau)},
\end{equation}
where $B$ is the batch size and $\tau$ represents the temperature parameter. 
This loss is designed to promote the cosine similarity between representations of different modalities originating from the same sample. Simultaneously, it pushes the representations of different samples away from each other.
Following the same principles, the loss for modality $b$ can be calculated as $\ell(z_b, z_a)$.
The final training loss can be defined as:
\begin{equation}
    \ell_{clip}(z_a, z_b) = \frac{1}{2} (\ell(z_a, z_b) + \ell(z_b, z_a)).
\end{equation}
In supervised LW-FedMML, the fused representations of $z_a$ and $z_b$ are passed through a prediction head $H_\text{sup}$ for making predictions, as shown in Fig~\ref{fig:sup_lwmml}. 
The cross-entropy loss is calculated between the prediction $\hat{y}$ and the ground truth label $y$ for training.

\begin{table*}[tb]
\centering
\caption{Comparison on how the layers are managed across different training approaches.}
\label{tab:comparison_characteristics}
\begin{tabular}{l|ccc|cccc}
\toprule
\textbf{}            & \textbf{}         & \textbf{}                                                           & \textbf{}                                                          & \multicolumn{4}{c}{\textbf{At Stage \textit{$\mathbf{s \in [1,S]}$}}}                                                                                                                                                                                                      \\ \cmidrule{5-8} 
\textbf{}            & \textbf{Training} & \textbf{\begin{tabular}[c]{@{}c@{}}Number of\\ Stages\end{tabular}} & \textbf{\begin{tabular}[c]{@{}c@{}}Starting\\ Layers\end{tabular}} & \textbf{\begin{tabular}[c]{@{}c@{}}Existing\\ Layers\end{tabular}} & \textbf{\begin{tabular}[c]{@{}c@{}}Added \\ Layer\end{tabular}} & \textbf{\begin{tabular}[c]{@{}c@{}}Frozen\\ Layers\end{tabular}} & \textbf{\begin{tabular}[c]{@{}c@{}}Active\\ Layers\end{tabular}} \\ \midrule
\textbf{FedMML}      & end-to-end        & --                                                                  & $L^1$ to $L^S$                                                     & $L^1$ to $L^S$                                                     & --                                                              & --                                                               & $L^1$ to $L^S$                                                   \\
\textbf{Prog-FedMML} & progressive       & $S$                                                                 & --                                                                 & $L^1$ to $L^s$                                                     & $L^s$                                                           & --                                                               & $L^1$ to $L^s$                                                   \\
\textbf{LW-FedMML}   & layer-wise        & $S$                                                                 & --                                                                 & $L^1$ to $L^s$                                                     & $L^s$                                                           & $L^1$ to $L^{(s-1)}$                                               & $L^s$                                                            \\ \bottomrule
\end{tabular}
\end{table*}

FL clients perform the aforementioned steps as the local training process.
The detailed procedure of LW-FedMML is shown in Algorithms~\ref{alg:server} and \ref{alg:client}. 
Algorithm~\ref{alg:server} describes the server-side execution, responsible for determining the stage $s$ and initializing new layers $L_a^s$ and $L_b^s$ for each modality, as well as completing the aggregation process.
On the other hand, the clients execute Algorithm~\ref{alg:client}, where they train the active layers $L_a^s$ and $L_b^s$ along with the projection heads $H_a$ and $H_b$.
For the supervised setting, clients also train the prediction head $H_\text{sup}$ and exchange it with the server. 
In Appendix~\ref{sec:sup_lwfedmml_algo}, we describe the detailed procedure for supervised LW-FedMML using Algorithms~\ref{alg:server_sup} and \ref{alg:client_sup}.
By training only the active layers in each stage, memory and computational resource requirements for clients are significantly reduced.
This enables resource-limited client devices to actively participate in the FL process, allowing to take advantage of their private data for model training.
Furthermore, only the active layers are required to exchange between the clients and the server. 
Consequently, both upload and download costs for model exchange are reduced, minimizing FL communication bottlenecks.

\subsection{Prog-FedMML}

Inspired by \cite{wang2022progfed}, we also explore the progressive training approach for federated multimodal learning, denoted as \mbox{Prog-FedMML}. 
Similar to the layer-wise approach, progressive training begins with an initial encoder $F^0$ without any layers. 
A new layer $L^s$ is sequentially added at each stage $s$.
However, unlike the layer-wise approach, all existing layers in $F^s$ are trained during each stage $s$.
Specifically for multimodal learning, Prog-FedMML trains layers $[L_a^1, \dots, L_a^{s}]$ and $[L_b^1, \dots, L_b^{s}]$ within encoders $F_a^s$ and $F_b^s$ during stage $s$.
Progressive training, as demonstrated in \cite{wang2022progfed} and \cite{tun2024lw}, has the potential to outperform end-to-end training.
This advantage likely stems from training and increasing the model depth step-by-step, which may reduce the risk of vanishing or exploding gradients and allow earlier layers to receive more effective gradient updates.
This phenomenon is examined further in Section~\ref{sec:gradient}.

\mbox{Prog-FedMML} is more resource-intensive compared to \mbox{LW-FedMML}, 
as it requires training all layers added in prior stages rather than freezing them.
The computational and communication resource requirements of Prog-FedMML for an FL training round gradually increase and eventually match those of end-to-end training by the final stage (i.e., when $s=S$).
This similarity in resource demands could lead to the same drawbacks as end-to-end training, particularly in excluding clients with limited resources from the FL process, thereby losing valuable data for model training.
We compare the different characteristics of end-to-end, progressive, and layer-wise training approaches for an encoder $F$ in Table~\ref{tab:comparison_characteristics}, highlighting how the layers are managed and utilized at each stage.

\begin{figure}[tb]
    \centering
    \includegraphics[width=0.9\linewidth]{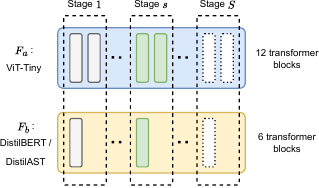}
    \caption{For training on the COCO dataset, we attach two transformer blocks to the ViT-Tiny encoder and one transformer block to the DistilBERT encoder at each stage. Similarly, for the ADVANCE dataset, we attach two transformer blocks to the ViT-Tiny encoder and one transformer block to the DistilAST encoder at each stage.}
    \label{fig:asymmetric_layer}
\end{figure}

\section{Experiment}

\subsection{Experimental Setup}

\emph{\textbf{Dataset.}} We conduct our experiments for multimodal learning in two settings: an instance discrimination-based setting and a supervised setting with labeled data. 
We use the COCO~\cite{lin2014microsoft} dataset for the instance discrimination-based setting.
This dataset contains 118,287 images for training and 5,000 images for testing, with each image paired with at least five captions.
We evaluate the performance on text-to-image and image-to-text retrieval tasks.
For the supervised setting, we use the ADVANCE~\cite{hu2020cross} dataset for audiovisual aerial scene recognition. 
This dataset consists of 5,075 aerial images accompanied by their respective audio recordings, categorized into 13 different scene classes. 
The underlying premise is that geographic locations are associated with certain sound events, and utilizing both aerial images and sound can lead to more accurate predictions~\cite{hu2020cross}.
We allocate 20\% of the ADVANCE dataset for testing.

\emph{\textbf{Model.}} We choose ViT-Tiny~\cite{dosovitskiy2020image} for the image modality, DistilBERT~\cite{sanh2019distilbert} for the text modality, and a distilled version of the audio spectrogram transformer~\cite{gong2021ast}, which we denote as DistilAST, for the audio modality.
We utilize pre-trained weights from the HuggingFace Transformers library~\cite{wolf2019huggingface} for these models. 
It is important to highlight that these pre-trained weights are derived from training separately on unimodal data, and not on multimodal data or tasks.
ViT-Tiny is paired with DistilBERT for the COCO dataset, and with DistilAST for the ADVANCE dataset.
The projection head $H$ is implemented as a 3-layer MLP for all modalities, with the dimension of hidden layers set to 4096 and the output dimension set to 256.
For the supervised setting, the prediction head $H_\text{sup}$ is a 2-layer MLP with hidden layer dimensions of 256.

\begin{table*}[tb]
\centering
\caption{Distribution of FL communication rounds to each stage of layer-wise training on the COCO dataset.}
\label{tab:num_rounds_distribution}
\begin{tabular}{l|cccccc|c|cccc|cccc}
\toprule
                               & \multicolumn{6}{c|}{\textbf{Rounds $\pmb{R^s}$ for Stage $\pmb{s}$}}                    & \multirow{2}{*}{\textbf{\begin{tabular}[c]{@{}c@{}}Total Number\\of  Rounds\end{tabular}}} & \multicolumn{4}{c|}{\textbf{Text-to-Image}}                 & \multicolumn{4}{c}{\textbf{Image-to-Text}}                  \\
\multicolumn{1}{c|}{\textbf{}} & \textbf{1} & \textbf{2} & \textbf{3} & \textbf{4} & \textbf{5} & \textbf{6} &                                                                                  & \textbf{R@1} & \textbf{R@5} & \textbf{R@10} & \textbf{R@50} & \textbf{R@1} & \textbf{R@5} & \textbf{R@10} & \textbf{R@50} \\ \midrule
\textbf{LW-FedMML}             & 15         & 15         & 15         & 15         & 15         & 15         & 90                                                                               & 21.26        & 48.54        & 62.21         & 89.30         & 30.48        & 59.32        & 71.58         & 92.50         \\
\textbf{LW-FedMML}             & 0          & 0          & 0          & 10         & 30         & 50         & 90                                                                               & \textbf{25.22}        & \textbf{53.82}        & \textbf{66.92}         & \textbf{91.30}         & \textbf{35.52}        & \textbf{65.78}        & \textbf{76.88}         & \textbf{94.20}         \\ \bottomrule
\end{tabular}
\end{table*}

\emph{\textbf{Training settings.}} For layer-wise training, we break down the training process into six stages.
In this context, the term ``layer" refers to a transformer block or a model portion that comprises multiple layers.
The ViT-Tiny architecture consists of 12 transformer blocks, while both DistilBERT and DistilAST contain six transformer blocks.
For the COCO dataset, we attach two transformer blocks to the ViT-Tiny encoder and one transformer block to the DistilBERT encoder at each training stage, as illustrated in Fig.~\ref{fig:asymmetric_layer}. 
Similarly, for the ADVANCE dataset, we attach two transformer blocks to the ViT-Tiny encoder and one transformer block to the DistilAST encoder at each stage, following the same training procedure.
We set the number of FL clients to 10, with each client performing local training for three epochs ($E=3$).
The total number of FL communication rounds $R$ is set to 90.

\begin{table}[tb]
\centering
\caption{Learning rates for encoders of different modalities.}
\label{tab:lr}
\begin{tabular}{l|c|cccccc}
\toprule
\multicolumn{1}{c|}{\multirow{2}{*}{\textbf{Dataset}}}             & \multirow{2}{*}{\textbf{Modality}} & \multicolumn{6}{c}{\textbf{Stage $\pmb{s}$}}                                \\
\textbf{}                         &                                    & \textbf{1} & \textbf{2} & \textbf{3} & \textbf{4} & \textbf{5} & \textbf{6} \\ \midrule
\multirow{2}{*}{\textbf{COCO}}    & \textbf{Image}                     & 1e-7       & 1e-7       & 1e-7       & 5e-6       & 3e-5       & 5e-5       \\
                                  & \textbf{Text}                      & 1e-7       & 1e-7       & 1e-7       & 1e-5       & 3e-5       & 5e-5       \\ \midrule
\multirow{2}{*}{\textbf{ADVANCE}} & \textbf{Image}                     & 1e-7       & 1e-7       & 1e-7       & 5e-7       & 1e-6       & 5e-6       \\
                                  & \textbf{Audio}                     & 1e-5       & 1e-5       & 1e-5       & 1e-5       & 1e-5       & 3e-5       \\ \bottomrule
\end{tabular}
\end{table}

We uniformly split the data among the clients for the COCO dataset. 
We set the batch size to 256 and use $\tau=0.05$ in Eq.~\ref{eq:infonce}.
For LW-FedMML and Prog-FedMML, we use the learning rates listed in Table~\ref{tab:lr} for the image and text encoders at each stage. 
These values are guided by the transfer learning principles, where pre-trained early layers capture generic features and require minimal fine-tuning, while later layers become more task-specific.
As a result, we use lower learning rates for the early stages and gradually increase them for the later stages.
The learning rate for the projection heads of both modalities is set to 1e-4.
For the conventional FedMML, the learning rates for the image and text encoders are set to 1e-7 and 1e-5, respectively, while the projection heads have a learning rate of 1e-4.
Unless otherwise stated, we use the COCO dataset for our experiments.

As for the ADVANCE dataset, we use the Dirichlet distribution~\cite{ferguson1973bayesian} with the concentration parameter $\beta=0.5$ to split the data among clients based on the class labels.  
The batch size is set to 16.
For LW-FedMML and Prog-FedMML, the learning rates are set according to Table~\ref{tab:lr}.
For FedMML, the learning rates for the image and audio encoders are set to 1e-7 and 1e-5, respectively. 
The projection heads and the prediction heads have a base learning rate of 1e-4 for all approaches.
The learning rate is linearly scaled as \textit{$\text{base learning rate} \times \text{batch size} / \textit{256}$} \cite{goyal2017accurate, krizhevsky2014one}.
By default, we use concatenation as the fusion approach for the ADVANCE dataset.

\subsection{Instance Discrimination-based Setting}

Given a limited number of FL rounds (i.e., $R=90$), we first determine how to allocate them to different stages for layer-wise training.
As shown in Table~\ref{tab:num_rounds_distribution}, we experiment with two strategies: uniformly distributing the FL rounds across all stages and allocating more rounds to the later stages. 
Prior work~\cite{tun2024lw} indicates that allocating the FL rounds uniformly or more to earlier stages is effective in the absence of pre-trained weights.
However, our findings in Table~\ref{tab:num_rounds_distribution} demonstrate that  allocating more rounds to the later stages achieves superior performance compared to uniform distribution when using pretrained weights.
Therefore, we adopt this allocation strategy in our subsequent experiments.
For Prog-FedMML, we use uniform allocation unless otherwise stated.

In Table~\ref{tab:clip_comparison}, we compare the performance (Recall@K) of different approaches and the relative resource requirements for a single client.
We measure the maximum memory required during training, while FLOPs and communication costs are measured in total across all local training epochs and FL rounds.
For FLOPs calculation, we follow the same strategy as in \cite{tun2024lw}, accounting for only one input sample rather than the entire batch.
As the number of samples increases, the FLOPs would also increase drastically.

Table~\ref{tab:clip_comparison} demonstrates that \mbox{LW-FedMML} achieves comparable performance to end-to-end FedMML, with a minor drop of $\approx 1\%$ in average recall for text-to-image and image-to-text retrieval tasks, while significantly reducing resource requirements.
Specifically, \mbox{LW-FedMML} shows a $2.7\times$ reduction in memory usage, a $2.4\times$ decrease in FLOPs requirements, and $2.3 \times$ lower communication costs.

\begin{figure*}[tb]
    \centering
    \subfloat[Memory.]{\includegraphics[width=0.3\linewidth]{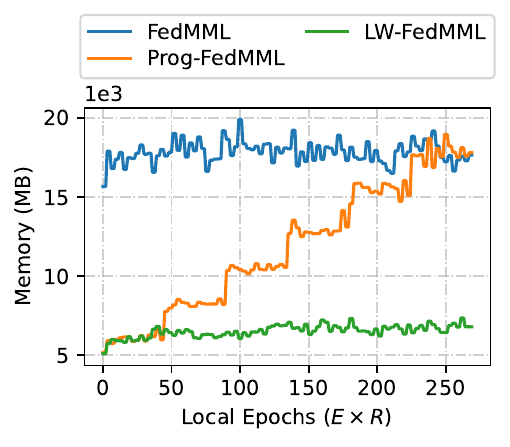} \label{subfig:mem_clip}}
    \subfloat[FLOPs.]{\includegraphics[width=0.3\linewidth]{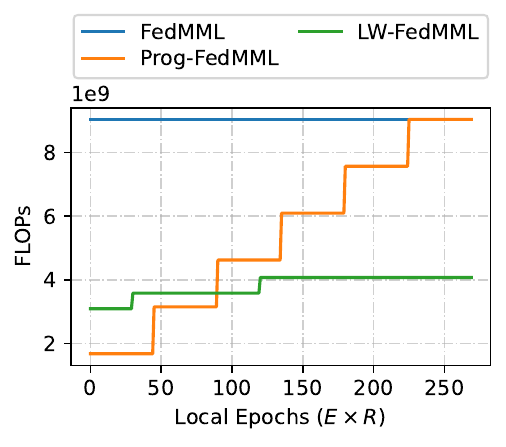} \label{subfig:flops_clip}}
    \subfloat[Communication.]{\includegraphics[width=0.3\linewidth]{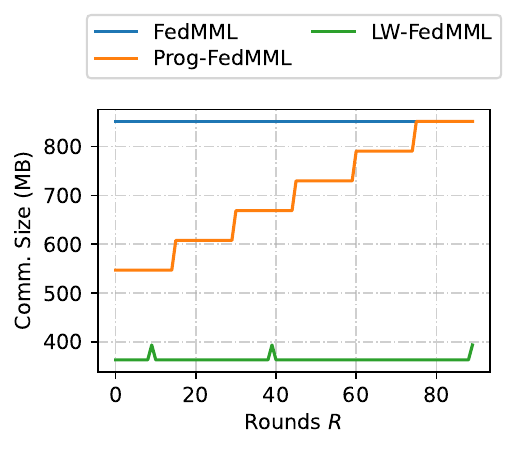} \label{subfig:comm_cost_clip}}
	\caption{Computational and communication resources required for a client on the COCO dataset. (a) Memory requirement. (b) FLOPs consumption. (c) Communication cost.}
	\label{fig:cost_clip}
\end{figure*}

Prog-FedMML achieves the best performance, surpassing end-to-end FedMML, which aligns with findings from prior studies \cite{wang2022progfed, tun2024lw}. 
However, the resource requirements of \mbox{Prog-FedMML} are significantly higher than those of \mbox{LW-FedMML}, closely matching those of end-to-end FedMML in terms of memory and communication costs.
This could present similar challenges for \mbox{Prog-FedMML} as those faced by end-to-end FedMML, where many clients may fail to meet the resource requirements for participation, thereby preventing their data from being used in model training.
Fig.~\ref{fig:cost_clip} plots the memory usage, FLOPs, and communication costs during the training process for a client.
Across all metrics, LW-FedMML consistently shows significantly lower resource requirements compared to FedMML.
In contrast, resource demands  for Prog-FedMML gradually increase throughout the training, eventually matching those of FedMML.

\begin{table*}[tp]
\centering
\caption{Comparison between FedMML (end-to-end), Prog-FedMML (progressive), and LW-FedMML (layer-wise) for text-to-image and image-to-text retrieval tasks on the COCO dataset. We also compare the relative resource requirements of each approach for a client.}
\label{tab:clip_comparison}
\begin{tabular}{l|ccc|clll|lccc}
\toprule
                               & \multicolumn{3}{c|}{\textbf{Resource Requirements}} & \multicolumn{4}{c|}{\textbf{Text-to-Image}}                                                                              & \multicolumn{4}{c}{\textbf{Image-to-Text}}                                      \\
\multicolumn{1}{c|}{\textbf{}} & \textbf{Memory}  & \textbf{FLOPs}  & \textbf{Comm.} & \textbf{R@1} & \multicolumn{1}{c}{\textbf{R@5}} & \multicolumn{1}{c}{\textbf{R@10}} & \multicolumn{1}{c|}{\textbf{R@50}} & \multicolumn{1}{c}{\textbf{R@1}} & \textbf{R@5} & \textbf{R@10} & \textbf{R@50} \\ \midrule
\textbf{FedMML}               & 1.00$\times$     & 1.00$\times$    & 1.00$\times$   & 26.31        & 55.06                            & 67.91                             & 91.58                              & 37.48                            & 66.00        & 78.12         & 94.52         \\
\textbf{Prog-FedMML}           & 0.95$\times$     & 0.59$\times$    & 0.82$\times$   & \textbf{27.42}        & \textbf{56.14}                            & \textbf{69.38}                             & \textbf{92.77}                              & \textbf{39.02}                            & \textbf{68.32}        & \textbf{79.10}         & \textbf{95.54}         \\
\textbf{LW-FedMML}             & \textbf{0.37}$\times$     & \textbf{0.42}$\times$    & \textbf{0.43}$\times$   & 25.22        & 53.82                            & 66.92                             & 91.30                              & 35.52                            & 65.78        & 76.88         & 94.20         \\ \bottomrule
\end{tabular}
\end{table*}

\begin{figure}[tb]
    \centering
    \subfloat[FedMML]{\includegraphics[width=0.487\linewidth]{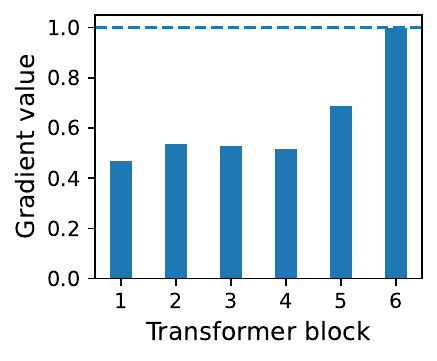} \label{subfig:ete_grad}} 
    \subfloat[Prog-FedMML]{\includegraphics[width=0.487\linewidth]{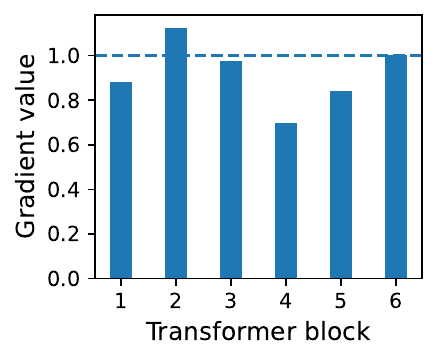} \label{subfig:prog_grad}} 
	\caption{Relative gradient values of linear layers corresponding to the attention output in DistilBERT. (a) For FedMML, the gradient value diminishes as it backpropagates to earlier layers. (b) Conversely, Prog-FedMML maintains improved gradient flow, allowing for more effective updates.}
	\label{fig:track_grad}
\end{figure}

\subsection{Tracking Gradients}
\label{sec:gradient}

We investigate the underlying factors that might contribute to the superior performance of progressive training compared to conventional end-to-end training.
Using the text encoder (i.e., DistilBERT) with six transformer blocks, we examine the gradient values of linear layers corresponding to the attention output in each transformer block. 
We use the same batch of samples for measuring the gradient values. 
For \mbox{FedMML}, we measure the gradients before the training when errors and gradient updates are expected to be the largest.
For \mbox{Prog-FedMML}, we measure the gradient value of each layer at the first round of its corresponding stage.
The gradient values relative to the layer in the last transformer block are illustrated in Fig.~\ref{fig:track_grad}.
For FedMML, the gradient value diminishes as it backpropagates to earlier layers. 
This diminishing pattern is not observed in Prog-FedMML.
Although not yet thoroughly investigated, increasing the model depth step-by-step may enable earlier layers to receive more effective gradient updates by mitigating small updates caused by vanishing gradients or large divergences due to exploding gradients.
We believe this is a key factor contributing to the superior performance of progressive training compared to conventional end-to-end training.

\subsection{Eligible Clients}
\label{sec:eligible_clients}

Due to the resource-constrained nature of FL clients, not all clients may be eligible to perform end-to-end or progressive training, which have high resource demands. 
This means that only a subset out of all available clients can contribute to \mbox{FedMML} and Prog-FedMML.
Consequently, the valuable data on these low-resource clients cannot be used, adversely affecting overall performance.
Table~\ref{tab:valid_client} compares this scenario, showing that as the number of clients eligible for training decreases, the performance of FedMML and \mbox{Prog-FedMML} also declines.
In contrast, LW-FedMML, with its lower resource requirements, has the potential to allow these clients to participate in the training.
This inclusion enables the use of data from low-resource devices, thereby improving the overall effectiveness of the training process.

\begin{table*}[tb]
\centering
\caption{Performance comparison on the COCO dataset, assuming that only a portion of clients can meet the resource requirements for end-to-end training (FedMML) and progressive training (Prog-FedMML) approaches.}
\label{tab:valid_client}
\begin{tabular}{l|ccc|c|cccc|cccc}
\toprule
                                                           & \multicolumn{3}{c|}{\textbf{Resource Requirements}}                                                                                                                                                               & \multirow{2}{*}{\textbf{\begin{tabular}[c]{@{}c@{}}Eligible\\ Clients\end{tabular}}} & \multicolumn{4}{c|}{\textbf{Text-to-Image}}                 & \multicolumn{4}{c}{\textbf{Image-to-Text}}                  \\
\textbf{}                                                  & \textbf{\begin{tabular}[c]{@{}c@{}}Memory\\ (MB)\end{tabular}} & \textbf{\begin{tabular}[c]{@{}c@{}}FLOPs\\ $\bm{(\times 10^{12})}$\end{tabular}} & \textbf{\begin{tabular}[c]{@{}c@{}}Comm.\\ (MB)\end{tabular}} &                                                                                      & \textbf{R@1} & \textbf{R@5} & \textbf{R@10} & \textbf{R@50} & \textbf{R@1} & \textbf{R@5} & \textbf{R@10} & \textbf{R@50} \\ \midrule
\multirow{2}{*}{\textbf{FedMML}}                           & \multirow{2}{*}{19913}                                         & \multirow{2}{*}{2.44}                                                            & \multirow{2}{*}{76613}                                        & 5/10 (50\%)                                                                          & 22.48        & 49.03        & 62.47         & 88.50         & 30.02        & 59.44        & 72.30         & 92.10         \\
                                                           &                                                                &                                                                                  &                                                               & 3/10 (30\%)                                                                          & 18.45        & 43.98        & 57.18         & 84.66         & 25.30        & 52.48        & 66.34         & 90.14         \\ \midrule
\multicolumn{1}{c|}{\multirow{2}{*}{\textbf{Prog-FedMML}}} & \multirow{2}{*}{18969}                                         & \multirow{2}{*}{1.45}                                                            & \multirow{2}{*}{62918}                                        & 5/10 (50\%)                                                                          & 23.89        & 52.22        & 65.27         & 90.94         & 34.38        & 63.08        & 74.78         & 93.70         \\
\multicolumn{1}{c|}{}                                      &                                                                &                                                                                  &                                                               & 3/10 (30\%)                                                                          & 19.48        & 46.08        & 59.82         & 87.24         & 26.76        & 56.82        & 69.02         & 91.44         \\ \midrule
\textbf{LW-FedMML}                                         & \textbf{7371}                                                           & \textbf{1.03}                                                                            & \textbf{32794}                                                         & 10/10 (100\%)                                                                        & \textbf{25.22}        & \textbf{53.82}        & \textbf{66.92}         & \textbf{91.30}         & \textbf{35.52}        & \textbf{65.78}        & \textbf{76.88}         & \textbf{94.20}         \\ \bottomrule
\end{tabular}
\end{table*}

\begin{figure*}[tb]
    \centering
    \subfloat[Memory usage.]{\includegraphics[width=0.3\linewidth]{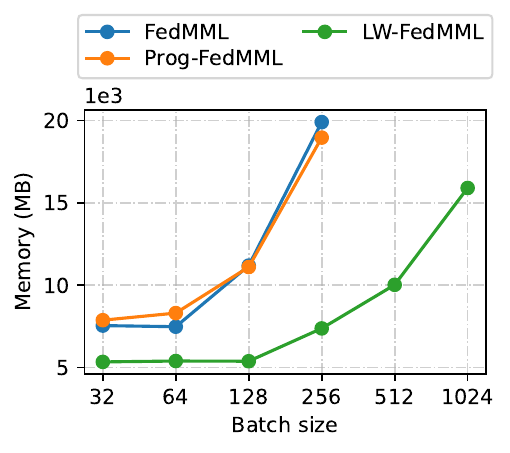} \label{subfig:batch_mem}}
    \subfloat[Text-to-Image.]{\includegraphics[width=0.3\linewidth]{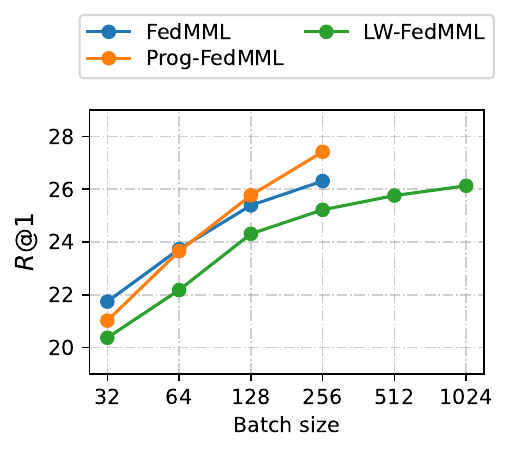} \label{subfig:batch_r1_tti}}
    \subfloat[Image-to-Text.]{\includegraphics[width=0.3\linewidth]{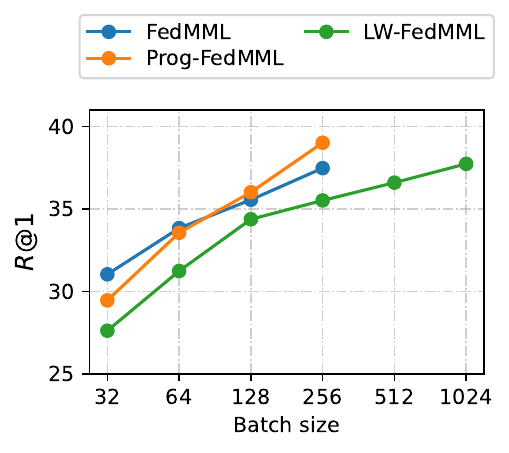} \label{subfig:batch_r1_itt}}
	\caption{Comparison with different batch sizes on the COCO dataset.}
	\label{fig:batch_size}
\end{figure*}

\begin{table*}[tb]
\centering
\caption{Time and energy consumption for a client.}
\label{tab:time_energy}
\begin{tabular}{l|cc|cc|cc}
\toprule
\textbf{Number of Samples} & \multicolumn{2}{c|}{\textbf{1000}}       & \multicolumn{2}{c|}{\textbf{10,000}}     & \multicolumn{2}{c}{\textbf{100,000}}     \\
                           & \textbf{Time (s)} & \textbf{Energy (Wh)} & \textbf{Time (s)} & \textbf{Energy (Wh)} & \textbf{Time (s)} & \textbf{Energy (Wh)} \\ \midrule
\textbf{FedMML}            & 326            & 23                 & 2080           & 144                 & 19865          & 1380                \\
\textbf{Prog-FedMML}       & 312            & 21                 & 1963           & 136                 & 18490          & 1284                \\
\textbf{LW-FedMML}         & \textbf{249}   & \textbf{17}        & \textbf{1107}  & \textbf{77}        & \textbf{9487}  & \textbf{659}       \\ \bottomrule
\end{tabular}
\end{table*}

\subsection{Batch Size}

Even though a smaller batch size can reduce memory usage, instance discrimination-based approaches benefit from larger batch sizes due to the increased number of samples.
We compare the maximum memory usage of each training approach with different batch sizes in Fig.~\ref{subfig:batch_mem}, while  Figs.~\ref{subfig:batch_r1_tti} and \ref{subfig:batch_r1_itt} compare their performance.
The results indicate that smaller batch sizes lead to reduced performance across all methods.
Fig.~\ref{subfig:batch_mem} shows a sharp increase in memory requirements for end-to-end FedMML and \mbox{Prog-FedMML} even as the batch size grows from 32 to 256. 
In contrast, the memory requirements for LW-FedMML remain relatively low across the same batch sizes. 
Fig.~\ref{fig:batch_size} also indicates that \mbox{LW-FedMML} with a batch size of 1024 uses less memory than end-to-end FedMML with a batch size of 256, while achieving the same level of performance.
Moreover, increasing the batch size in \mbox{LW-FedMML} only impacts the memory usage, allowing it to maintain its advantage of lower communication costs.

\subsection{Time and Energy Consumption}

In Table~\ref{tab:time_energy}, we compare the time and energy consumption of a client based on the size of its local dataset.
It is important to emphasize that we only conduct a simplified  investigation, as time and energy consumption can vary significantly depending on the hardware used. 
For this study, we simply measure the total training time of a client, assuming that the client device operates in maximum power mode for all approaches during training.
Specifically, we used a V100 GPU (32 GB), with a maximum power consumption of 250 watts.
Energy consumption can then be calculated as \textit{$\text{energy} = \text{training time} \times \text{power}$}.
The results show that as the number of data samples within the client's dataset increases, the time and energy consumption gap between \mbox{LW-FedMML} and FedMML becomes more significant.

\begin{table*}[tb]
\centering
\caption{Comparison between FedMML (end-to-end), Prog-FedMML (progressive), and LW-FedMML (layer-wise) for audiovisual aerial scene recognition on the ADVANCE dataset. We also compare the relative resource requirements of each approach for a client.}
\label{tab:sup_comparison}
\begin{tabular}{lccccccc}
\toprule
\multicolumn{1}{l|}{\textbf{}}   & \multicolumn{3}{c|}{\textbf{Resource Requirements}}                                                 &                &                &                &                \\
\multicolumn{1}{l|}{}            & \textbf{Memory}                & \textbf{FLOPs}                 & \multicolumn{1}{c|}{\textbf{Comm.}}                 & \textbf{Precision}      & \textbf{Recall}         & \textbf{F1}             & \textbf{Accuracy}       \\ \midrule
                                 & \multicolumn{7}{c}{\textbf{Concat}}                                                                                                                                     \\ \midrule
\multicolumn{1}{l|}{\textbf{FedMML}}      & 1.00$\times$          & 1.00$\times$          & \multicolumn{1}{c|}{1.00$\times$}          & \textbf{75.91} & \textbf{73.79} & \textbf{74.78} & 95.46          \\
\multicolumn{1}{l|}{\textbf{Prog-FedMML}} & 1.00$\times$          & 0.91$\times$          & \multicolumn{1}{c|}{0.95$\times$}          & 74.79          & 72.91          & 73.80          & \textbf{95.76} \\
\multicolumn{1}{l|}{\textbf{LW-FedMML}}   & \textbf{0.63$\times$} & \textbf{0.41$\times$} & \multicolumn{1}{c|}{\textbf{0.54$\times$}} & 74.19          & 72.17          & 73.11          & 95.17          \\ \midrule
                                 & \multicolumn{7}{c}{\textbf{Concat + Add}}                                                                                                                               \\ \midrule
\multicolumn{1}{l|}{\textbf{FedMML}}      & 1.00$\times$          & 1.00$\times$          & \multicolumn{1}{c|}{1.00$\times$}          & 73.18          & 71.30          & 72.16          & 94.58          \\
\multicolumn{1}{l|}{\textbf{Prog-FedMML}} & 1.00$\times$          & 0.91$\times$          & \multicolumn{1}{c|}{0.95$\times$}          & \textbf{81.12} & \textbf{79.64} & \textbf{80.33} & \textbf{96.68} \\
\multicolumn{1}{l|}{\textbf{LW-FedMML}}   & \textbf{0.61$\times$} & \textbf{0.41$\times$} & \multicolumn{1}{c|}{\textbf{0.54$\times$}} & 74.48          & 72.53          & 73.44          & 95.02          \\ \midrule
                                 & \multicolumn{7}{c}{\textbf{Concat + Self-Attention \cite{torchmultimodal}}}                                                                                                                    \\ \midrule
\multicolumn{1}{l|}{\textbf{FedMML}}      & 1.00$\times$          & 1.00$\times$          & \multicolumn{1}{c|}{1.00$\times$}          & 71.85          & 69.70          & 70.71          & 94.98          \\
\multicolumn{1}{l|}{\textbf{Prog-FedMML}} & 1.00$\times$          & 0.91$\times$          & \multicolumn{1}{c|}{0.95$\times$}          & \textbf{76.35} & \textbf{74.45} & \textbf{75.34} & \textbf{95.56} \\
\multicolumn{1}{l|}{\textbf{LW-FedMML}}   & \textbf{0.60$\times$} & \textbf{0.41$\times$} & \multicolumn{1}{c|}{\textbf{0.54$\times$}} & 74.01          & 72.00          & 72.93          & 94.82          \\ \midrule
                                 & \multicolumn{7}{c}{\textbf{Concat + Co-Attention \cite{xiong2022unified}}}                                                                                                                      \\ \midrule
\multicolumn{1}{l|}{\textbf{FedMML}}      & 1.00$\times$          & 1.00$\times$          & \multicolumn{1}{c|}{1.00$\times$}          & 69.79          & 67.32          & 68.48          & 94.39          \\
\multicolumn{1}{l|}{\textbf{Prog-FedMML}} & 1.00$\times$          & 0.91$\times$          & \multicolumn{1}{c|}{0.95$\times$}          & 72.03          & 69.76          & 70.80          & \textbf{94.63} \\
\multicolumn{1}{l|}{\textbf{LW-FedMML}}   & \textbf{0.61$\times$} & \textbf{0.41$\times$} & \multicolumn{1}{c|}{\textbf{0.54$\times$}} & \textbf{73.62} & \textbf{71.41} & \textbf{72.40} & 94.43          \\ \bottomrule
\end{tabular}
\end{table*}

\begin{figure*}[tb]
    \centering
    \subfloat[Memory.]{\includegraphics[width=0.3\linewidth]{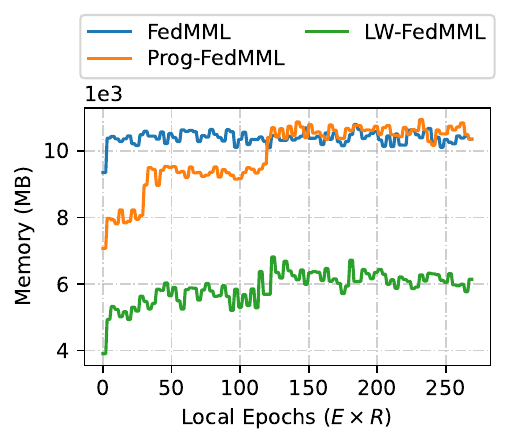} \label{subfig:mem_sup}}
    \subfloat[FLOPs.]{\includegraphics[width=0.3\linewidth]{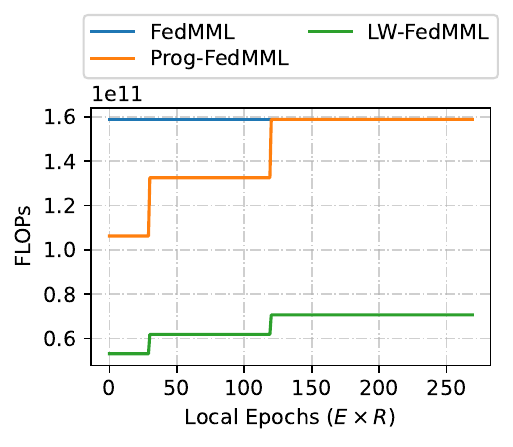} \label{subfig:flops_sup}}
    \subfloat[Communication.]{\includegraphics[width=0.3\linewidth]{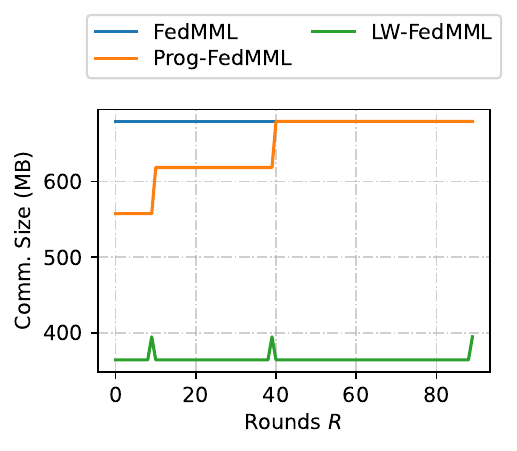} \label{subfig:comm_cost_sup}}
	\caption{Computational and communication resources required for a client when using the ADVANCE dataset with concatenation as the fusion technique. (a) Memory requirement. (b) FLOPs consumption. (c) Communication cost.}
	\label{fig:cost_sup}
\end{figure*}

\subsection{Supervised Setting}
\label{sec:sup_setting}

For the ADVANCE dataset, we allocate more FL rounds to later stages of 
\mbox{Prog-FedMML}, as we do for \mbox{LW-FedMML} in Table~\ref{tab:num_rounds_distribution}.
We use various fusion techniques, including concatenation, addition, self-attention~\cite{torchmultimodal}, and co-attention~\cite{xiong2022unified} to combine the image and audio projections for the prediction.
Table~\ref{tab:sup_comparison} compares the performance of different approaches and relative resource requirements for a single client.
We follow the same approach as in the instance discrimination-based setting for measuring resources.
The results show that Prog-FedMML achieves the best performance in terms of accuracy, while LW-FedMML either maintains competitive performance or outperforms FedMML across various metrics.
There is no significant variation in terms of relative resource requirements when using different fusion techniques.
Fig.~\ref{fig:cost_sup} plots the resource requirements of a client when using concatenation as the fusion approach. 
The plots indicate that LW-FedMML significantly lowers the resource requirements compared to FedMML, while \mbox{Prog-FedMML} exhibits similar resource demands as FedMML.

\begin{figure*}[tb]
    \centering
    \subfloat[$\beta=50$]{\includegraphics[width=0.235\linewidth]{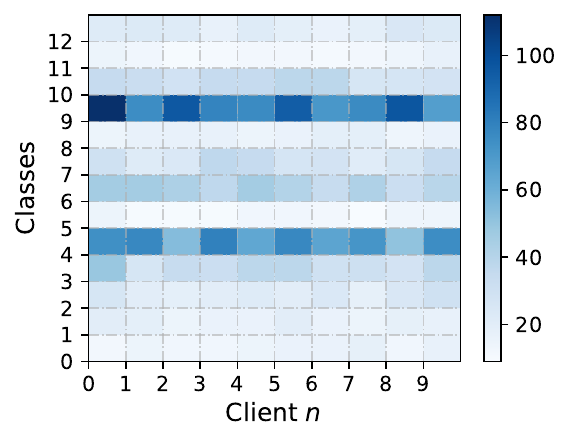} \label{subfig:beta_50}} 
    \subfloat[$\beta=5$]{\includegraphics[width=0.235\linewidth]{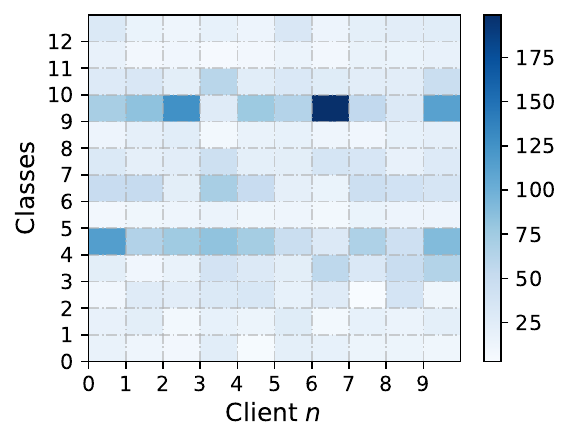} \label{subfig:beta_5}} 
    \subfloat[$\beta=0.5$]{\includegraphics[width=0.235\linewidth]{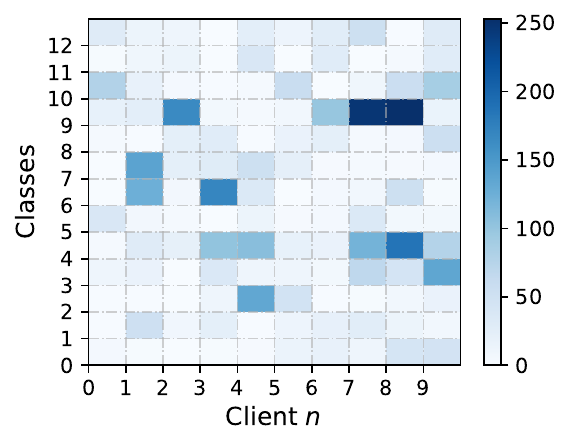} \label{subfig:beta_0.5}} 
    \subfloat[$\beta=0.05$]{\includegraphics[width=0.235\linewidth]{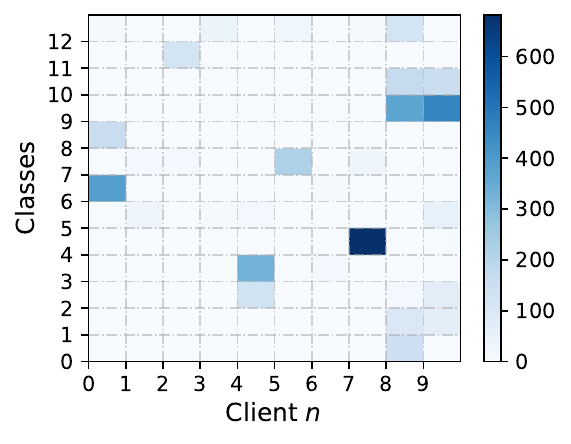} \label{subfig:beta_0.05}} 
	\caption{Data distribution among clients with different $\beta$ values for the ADVANCE dataset. A darker box represents a higher number of samples for a data class within a client.}
	\label{fig:betas_distribution}
\end{figure*}

\subsection{Data Heterogeneity}

Data heterogeneity is one of the main challenges in FL~\cite{zhao2018federated}.
Variations in device usage patterns, geographical locations, and other factors influence the data distribution among clients. 
To control the degree of data heterogeneity within the clients' local datasets, we adjust the $\beta$ parameter in the Dirichlet distribution.
A smaller $\beta$ value corresponds to a higher level of data heterogeneity, and vice versa.
Fig.~\ref{fig:betas_distribution} illustrates the data distribution among clients with different $\beta$ values for the ADVANCE dataset, and Table~\ref{tab:sup_data_hetero} shows their results. 
These results indicate that the performance of all approaches declines as the level of data heterogeneity increases. 
Nonetheless, both Prog-FedMML and LW-FedMML consistently maintain strong performance across all levels of data heterogeneity.


\begin{table}[tb]
\centering
\caption{Performance comparison across varying levels of data heterogeneity on the ADVANCE dataset.}
\label{tab:sup_data_hetero}
\begin{tabular}{lcccc}
\toprule
\multicolumn{1}{l|}{}                     & \textbf{Precision} & \textbf{Recall} & \textbf{F1}    & \textbf{Accuracy} \\ \midrule
                                          & \multicolumn{4}{c}{$\beta=50$}                                            \\ \midrule
\multicolumn{1}{l|}{\textbf{FedMML}}      & 78.80              & 77.04           & 77.84          & 95.91             \\
\multicolumn{1}{l|}{\textbf{Prog-FedMML}} & \textbf{80.78}     & \textbf{79.21}  & \textbf{79.95} & \textbf{96.49}    \\
\multicolumn{1}{l|}{\textbf{LW-FedMML}}   & 77.60              & 75.63           & 76.55          & 95.52             \\ \midrule
                                          & \multicolumn{4}{c}{$\beta=5$}                                             \\ \midrule
\multicolumn{1}{l|}{\textbf{FedMML}}      & 75.39              & 73.41           & 74.33          & 95.52             \\
\multicolumn{1}{l|}{\textbf{Prog-FedMML}} & \textbf{80.60}     & \textbf{79.10}  & \textbf{79.81} & \textbf{96.64}    \\
\multicolumn{1}{l|}{\textbf{LW-FedMML}}   & 77.24              & 75.39           & 76.24          & 95.42             \\ \midrule
                                          & \multicolumn{4}{c}{$\beta=0.5$}                                           \\ \midrule
\multicolumn{1}{l|}{\textbf{FedMML}}      & \textbf{75.91}     & \textbf{73.79}  & \textbf{74.78} & 95.46             \\
\multicolumn{1}{l|}{\textbf{Prog-FedMML}} & 74.79              & 72.91           & 73.80          & \textbf{95.76}    \\
\multicolumn{1}{l|}{\textbf{LW-FedMML}}   & 74.19              & 72.17           & 73.11          & 95.17             \\ \midrule
                                          & \multicolumn{4}{c}{$\beta=0.05$}                                          \\ \midrule
\multicolumn{1}{l|}{\textbf{FedMML}}      & 52.46              & 46.28           & 48.37          & 73.98             \\
\multicolumn{1}{l|}{\textbf{Prog-FedMML}} & 52.33              & 46.44           & 48.25          & \textbf{74.61}    \\
\multicolumn{1}{l|}{\textbf{LW-FedMML}}   & \textbf{54.96}     & \textbf{48.68}  & \textbf{50.74} & 74.12             \\ \bottomrule
\end{tabular}
\end{table}

\subsection{Client Participation Ratio}
\label{sec:client_participate}

In an FL environment with $N$ total clients, it is often impractical for all $N$ clients to simultaneously participate in every FL training round. 
Therefore, at each FL round, we randomly select a subset of clients to participate in the training process.
Using the ADVANCE dataset, we fixed the total number of clients to $N=50$ and set the number of participating clients $N_\text{p}$ to be 35, 25, and 15.
The total number of FL communication rounds $R$ is set to 300. 
For both \mbox{LW-FedMML} and \mbox{Prog-FedMML}, we allocate 50 FL rounds for each stage $s$.
The learning rate is set to 1e-4 for all modules in FedMML and Prog-FedMML.
For LW-FedMML, we set the learning rate of image and audio encoders to 1e-4, and to 1e-3 for the remaining modules.
The results in Table~\ref{tab:client_participate} demonstrate that LW-FedMML can maintain decent performance even when the client participation ratio is low, despite having significantly lower resource requirements compared to FedMML.
Meanwhile, Prog-FedMML achieves the best performance in most cases.

We also consider an experiment where each client is assigned a dropout probability, simulating instances where clients may leave the training process due to communication failures, bandwidth limitations or other operational constraints. 
In this setting, we set $N=50$ and examine two dropout probabilities: 25\% and 50\%. 
The results for this setting are shown in Table~\ref{tab:cdropoutrate}.


\begin{table}[tb]
\centering
\caption{Performance comparison on the ADVANCE dataset with different numbers of participating clients $N_\text{p}$ at each FL round. The total number of clients $N$ is  fixed at 50, while $N_\text{p}$ is set to 35, 25, and 15.}
\label{tab:client_participate}
\begin{tabular}{lcccc}
\toprule
\multicolumn{1}{l|}{}                     & \textbf{Precision} & \textbf{Recall} & \textbf{F1}    & \textbf{Accuracy} \\ \midrule
                                          & \multicolumn{4}{c}{$N_\text{p}/N=35/50$ $(70\%)$}                         \\ \midrule
\multicolumn{1}{l|}{\textbf{FedMML}}      & 72.89              & 70.42           & 71.54          & 94.29             \\
\multicolumn{1}{l|}{\textbf{Prog-FedMML}} & \textbf{77.21}     & \textbf{74.78}  & \textbf{75.81} & \textbf{94.83}    \\
\multicolumn{1}{l|}{\textbf{LW-FedMML}}   & 70.52              & 68.38           & 69.36          & 94.00             \\ \midrule
                                          & \multicolumn{4}{c}{$N_\text{p}/N=25/50$ $(50\%)$}                         \\ \midrule
\multicolumn{1}{l|}{\textbf{FedMML}}      & 72.50              & 70.12           & 71.20          & 94.19             \\
\multicolumn{1}{l|}{\textbf{Prog-FedMML}} & \textbf{76.43}     & \textbf{73.76}  & \textbf{74.90} & 94.11             \\
\multicolumn{1}{l|}{\textbf{LW-FedMML}}   & 72.40              & 70.16           & 71.18          & \textbf{95.27}    \\ \midrule
                                          & \multicolumn{4}{c}{$N_\text{p}/N=15/50$ $(25\%)$}                         \\ \midrule
\multicolumn{1}{l|}{\textbf{FedMML}}      & 73.80              & 71.50           & 72.55          & \textbf{94.68}    \\
\multicolumn{1}{l|}{\textbf{Prog-FedMML}} & \textbf{76.84}     & \textbf{73.86}  & \textbf{75.06} & 94.30             \\
\multicolumn{1}{l|}{\textbf{LW-FedMML}}   & 70.70              & 68.15           & 69.30          & 93.90             \\ \bottomrule
\end{tabular}
\end{table}

\begin{table}[tb]
\centering
\caption{Performance comparison on the ADVANCE dataset with varying client dropout rates.}
\label{tab:cdropoutrate}
\begin{tabular}{lcccc}
\toprule
\multicolumn{1}{l|}{}                     & \textbf{Precision} & \textbf{Recall} & \textbf{F1}    & \textbf{Accuracy} \\ \midrule
                                          & \multicolumn{4}{c}{dropout rate = 25\%}                                   \\ \midrule
\multicolumn{1}{l|}{\textbf{FedMML}}      & 72.37              & 69.85           & 70.99          & 94.29             \\
\multicolumn{1}{l|}{\textbf{Prog-FedMML}} & \textbf{77.08}     & \textbf{74.70}  & \textbf{75.72} & 94.79             \\
\multicolumn{1}{l|}{\textbf{LW-FedMML}}   & 72.53              & 70.35           & 71.33          & \textbf{95.07}    \\ \midrule
                                          & \multicolumn{4}{c}{dropout rate = 50\%}                                   \\ \midrule
\multicolumn{1}{l|}{\textbf{FedMML}}      & 75.10              & 72.83           & 73.86          & 94.68             \\
\multicolumn{1}{l|}{\textbf{Prog-FedMML}} & \textbf{78.52}     & \textbf{76.23}  & \textbf{77.22} & \textbf{95.02}    \\
\multicolumn{1}{l|}{\textbf{LW-FedMML}}   & 72.60              & 70.38           & 71.36          & 94.04             \\ \bottomrule
\end{tabular}
\end{table}

\begin{table*}[tb]
\centering
\caption{ResNet-18 as image encoder.}
\label{tab:sup_model_resnet}
\begin{tabular}{l|ccc|cccc}
\toprule
                     & \multicolumn{3}{c|}{\textbf{Resource Requirements}}                   &                    &                 &                &                   \\
                     & \textbf{Memory}       & \textbf{FLOPs}        & \textbf{Comm.}        & \textbf{Precision} & \textbf{Recall} & \textbf{F1}    & \textbf{Accuracy} \\ \midrule
\textbf{FedMML}      & 1.00$\times$          & 1.00$\times$          & 1.00$\times$          & \textbf{69.97}     & \textbf{67.69}  & \textbf{68.74} & \textbf{94.39}    \\
\textbf{Prog-FedMML} & 1.00$\times$          & 0.80$\times$          & 0.85$\times$          & 69.38              & 66.72           & 67.88          & 93.65             \\
\textbf{LW-FedMML}   & \textbf{0.66$\times$} & \textbf{0.41$\times$} & \textbf{0.57$\times$} & 69.56              & 67.13           & 68.21          & 93.55             \\ \bottomrule
\end{tabular}
\end{table*}

\begin{table*}[tb]
\centering
\caption{Training from scratch.}
\label{tab:sup_scratch}
\begin{tabular}{lcccccccc}
\toprule
\multicolumn{1}{l|}{}                     & \multicolumn{1}{c|}{}                    & \multicolumn{3}{c|}{\textbf{Resource Requirements}}                                        &                    &                 &                &                   \\
\multicolumn{1}{l|}{}                     & \multicolumn{1}{c|}{\textbf{Rounds $R$}} & \textbf{Memory}       & \textbf{FLOPs}        & \multicolumn{1}{c|}{\textbf{Comm.}}        & \textbf{Precision} & \textbf{Recall} & \textbf{F1}    & \textbf{Accuracy} \\ \midrule
                                          & \multicolumn{8}{c}{Same Number of Rounds}                                                                                                                                                                         \\ \midrule
\multicolumn{1}{l|}{\textbf{FedMML}}      & \multicolumn{1}{c|}{300}                 & 1.00$\times$          & 1.00$\times$          & \multicolumn{1}{c|}{1.00$\times$}          & 46.45              & 39.47           & 41.91          & \textbf{76.69}    \\
\multicolumn{1}{l|}{\textbf{Prog-FedMML}} & \multicolumn{1}{c|}{300}                 & 1.00$\times$          & 0.59$\times$          & \multicolumn{1}{c|}{0.78$\times$}          & \textbf{47.18}     & \textbf{39.90}  & \textbf{42.27} & 74.83             \\
\multicolumn{1}{l|}{\textbf{LW-FedMML}}   & \multicolumn{1}{c|}{300}                 & \textbf{0.71$\times$} & \textbf{0.31$\times$} & \multicolumn{1}{c|}{\textbf{0.54$\times$}} & 41.02              & 33.32           & 36.23          & 75.81             \\ \midrule
                                          & \multicolumn{8}{c}{Same Comm. Cost}                                                                                                                                                                               \\ \midrule
\multicolumn{1}{l|}{\textbf{FedMML}}      & \multicolumn{1}{c|}{161}                 & 1.00$\times$          & 1.00$\times$          & \multicolumn{1}{c|}{1.00$\times$}          & 44.90              & 35.92           & 38.53          & 68.73             \\
\multicolumn{1}{l|}{\textbf{Prog-FedMML}} & \multicolumn{1}{c|}{228}                 & 0.92$\times$          & 0.66$\times$          & \multicolumn{1}{c|}{1.00$\times$}          & \textbf{49.52}     & \textbf{41.11}  & \textbf{43.66} & 70.59             \\
\multicolumn{1}{l|}{\textbf{LW-FedMML}}   & \multicolumn{1}{c|}{300}                 & \textbf{0.71$\times$} & \textbf{0.57$\times$} & \multicolumn{1}{c|}{1.00$\times$}          & 41.02              & 33.32           & 36.23          & \textbf{75.81}    \\ \bottomrule
\end{tabular}
\end{table*}

\subsection{ResNet as Image Encoder}

When using model architectures other than transformers, appending new layers to the encoder at each stage can alter the encoder output dimensions.
To investigate this, we apply ResNet-18~\cite{he2016deep} as the image encoder and DistilAST as the audio encoder on the ADVANCE dataset, setting the number of stages to $S=4$.
As an additional ResNet block is introduced to the image encoder at each stage, the dimension of the output image representation changes.
Specifically, the output dimensions of the image encoder across the four stages are (64, 128, 256, 512), while the input dimension to the projection head remains fixed at 512.
Therefore, we introduce a linear layer to reshape the encoder output whenever there is a dimensionality mismatch.
This linear layer is discarded after its corresponding stage.
For DistilAST, we introduce two transformer blocks at each of the first two stages, followed by one transformer block at each of the last two stages, resulting in a total of six transformer blocks.

The results are shown in Table~\ref{tab:sup_model_resnet}.
For all approaches, the learning rates are set as follows: 1e-7 for the image encoder, \mbox{1e-5} for the audio encoder, and 1e-4 for all other modules. 
The total number of FL communication rounds is 90, with rounds for each stage allocated as (10, 20, 25, 35) for both \mbox{LW-FedMML} and Prog-FedMML.
Both Prog-FedMML and \mbox{LW-FedMML} demonstrate comparable performance to end-to-end FedMML, with LW-FedMML offering the additional advantage of significantly lower resource consumption.

\subsection{Training from Scratch}

Using a similar setup to Section~\ref{sec:client_participate}, we conduct the training without pre-trained weights.
Specifically, we use the setting where the number of participating clients in each round is set to $N_\text{p}=15$.
We also keep other settings except that for \mbox{LW-FedMML}, the learning rate for the image and audio encoders is set to 1e-3, and to 1e-4 for other modules.
Table~\ref{tab:sup_scratch} presents the results under two scenarios: one where the number of FL communication rounds is the same for all approaches, and another where the communication cost is the same.
Under the same communication cost, LW-FedMML achieves higher accuracy than FedMML while also using less memory and fewer FLOPs.
The results in Table~\ref{tab:sup_scratch} indicate that both \mbox{Prog-FedMML} and LW-FedMML can also operate without using the pre-trained weights.

\section{Conclusion}
\label{sec:conclusion}

Resource constraints are a significant challenge in the FL environment, mainly because FL clients are often edge devices with limited capabilities. 
This challenge is exacerbated in FL scenarios involving multimodal data due to the complexity of the models required for processing such data. 
Many clients cannot meet the resource requirements for conventional end-to-end training approaches in these scenarios, resulting in their valuable data being excluded from the training process.
To address this issue, we introduce LW-FedMML, a layer-wise training approach designed to significantly reduce the resource burden on clients. 
Our experiments demonstrate that \mbox{LW-FedMML} can compete with conventional end-to-end \mbox{FedMML} while achieving up to $2.7 \times$ lower memory usage, $2.4 \times$ FLOPs requirements, and $2.3 \times$ cheaper communication costs on the COCO dataset.
These findings highlight the potential of \mbox{LW-FedMML} to enable efficient and scalable multimodal learning in resource-constrained FL environments.
Additionally, we explore Prog-FedMML, a progressive training approach for federated multimodal learning.
While Prog-FedMML has higher resource requirements than \mbox{LW-FedMML}, it has the potential to outperform end-to-end \mbox{FedMML} in most cases.

\appendices
\section{Supervised LW-FedMML}
\label{sec:sup_lwfedmml_algo}

Algorithms~\ref{alg:server_sup} and \ref{alg:client_sup} describe the details for the supervised setting of \mbox{LW-FedMML}.
In the supervised setting, the local dataset $D^n$ on each client $n$ contains ground truth labels $y$. 
The prediction head $H_\text{sup}$ generates predictions $\hat{y}$ and the cross-entropy loss is calculated between $y$ and $\hat{y}$.
In addition to the active layers and the projection heads, the clients also exchange $H_\text{sup}$ with the server in the supervised setting.

\begin{figure*}[tb]
    \centering
    \includegraphics[width=\linewidth]{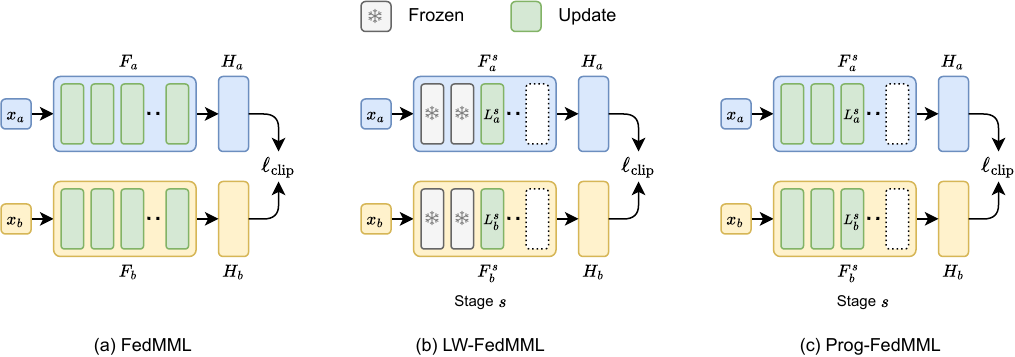}
    \caption{Comparison between: (a) FedMML (end-to-end), (b) Prog-FedMML (progressive), and (c) LW-FedMML (layer-wise) for the instance discrimination-based setting.}
    \label{fig:three_comparison}
\end{figure*}

\begin{algorithm}[tb]
\caption{Supervised LW-FedMML (\emph{Server-side})}
\label{alg:server_sup}

\textbf{Input:} encoder $F_a^0$ and projection head $H_a$ for modality $a$, encoder $F_b^0$ and projection head $H_b$ for modality $b$, prediction head $H_\text{sup}$, number of stages $S$, number of rounds $R^s$ where  $s \in [1, S]$ \\
\textbf{Output:} $F_a^S$, $F_b^S$, $H_a$, $H_b$, $H_\text{sup}$

\begin{algorithmic}[1]    
    \State \textbf{Server executes:}

    \For{stage $s=1,2,\dots, S$}
        \State Initialize new layers: $L_a^s$ and $L_b^s$
        \vspace{0.2em}
        \For{round $r=1,2,\dots, R^s$}
            \For{client $n=1,2,\dots,N$ in parallel}
                \vspace{0.2em}
                \State $L_a^{s,n}, L_b^{s,n}, H_a^n, H_b^n, H^n_\text{sup} \leftarrow $ 
                \vspace{0.2em}
                \item[] \hspace{10em} $\text{Train}(L_a^s, L_b^s, H_a, H_b, H_\text{sup})$
                \vspace{0.2em}
                \State $w^n = \frac{|D^n|}{|D|}$, where $D=\bigcup_{n=1}^{N} D^n$

            \EndFor

            \vspace{0.2em}
            \State $L_a^s \leftarrow \sum_{n=1}^N w^n L_a^{s,n}, \quad H_a \leftarrow \sum_{n=1}^N w^n H_a^n$
            \vspace{0.2em}
            \State $L_b^s \leftarrow \sum_{n=1}^N w^n L_b^{s,n}, \quad H_b \leftarrow \sum_{n=1}^N w^n H_b^n$
            \vspace{0.2em}
            \State $H_\text{sup} \leftarrow \sum_{n=1}^N w^n H^n_\text{sup}$
            
        \EndFor
        \State Distribute $L_a^s$ and $L_b^s$ to clients
        \vspace{0.2em}
        \State $F_a^s \leftarrow$ Attach $L_a^s$ to $F_a^{(s-1)}$
        \vspace{0.2em}
        \State $F_b^s \leftarrow$ Attach $L_b^s$ to $F_b^{(s-1)}$
        \vspace{0.2em}
    \EndFor
    \State return $F_a^S$, $F_b^S$, $H_a$, $H_b$, $H_\text{sup}$
    
\end{algorithmic}
\end{algorithm}

\begin{algorithm}[tb]
\caption{Supervised LW-FedMML (\emph{Client-side})}
\label{alg:client_sup}

\textbf{Input:} local dataset $D^n$, layers $[L_a^1, \dots, L_a^{(s-1)}]$ for modality~$a$, layers $[L_b^1, \dots, L_b^{(s-1)}]$ for modality $b$, number of local epochs~$E$  \\
\textbf{Output:} $L_a^{s,n}, L_b^{s,n}, H_a^n, H_b^n$, $H^n_\text{sup}$

\begin{algorithmic}[1]
    \State \textbf{Client executes:} $\text{Train}(L_a^s, L_b^s, H_a, H_b, H_\text{sup})$:
    \vspace{0.2em}
    \State $L_a^{s,n} \leftarrow L_a^s, \quad H_a^n \leftarrow H_a$
    \vspace{0.2em}
    \State $L_b^{s,n} \leftarrow L_b^s, \quad H_b^n \leftarrow H_b$
    \vspace{0.2em}
    \State $H^n_\text{sup} \leftarrow H_\text{sup}$
    \vspace{0.2em}
    \State $F_a^{s,n} \leftarrow$ Freeze $[L_a^1, \dots, L_a^{(s-1)}]$ and attach $L_a^{s,n}$
    \vspace{0.2em}
    \State $F_b^{s,n} \leftarrow$ Freeze $[L_b^1, \dots, L_b^{(s-1)}]$ and attach $L_b^{s,n}$
    \vspace{0.2em}
    \For{epoch $e=1,2,\dots,E$}
        \For{each batch $(x_a, x_b, y) \in D^n$}
            \vspace{0.2em}
            \State $z_a \leftarrow H_a^n(F_a^{s,n}(x_a))$
            \vspace{0.2em}
            \State $z_b \leftarrow H_b^n(F_b^{s,n}(x_b))$
            \vspace{0.2em}
            \State $\hat{z} \leftarrow$ Fuse $(z_a, z_b)$
            \vspace{0.2em}
            \State $\hat{y} \leftarrow H^n_\text{sup}(\hat{z})$
            \vspace{0.2em}
            \State $\mathcal{L} \leftarrow \ell_{\text{CE}}(y, \hat{y})$
            \vspace{0.2em}
            \State $L_a^{s,n}, L_b^{s,n}, H_a^n, H_b^n, H^n_\text{sup} \leftarrow$ Update with $\nabla\mathcal{L}$
            \vspace{0.2em}
        \EndFor
    \EndFor
    \State Upload $L_a^{s,n}, L_b^{s,n}, H_a^n, H_b^n, H^n_\text{sup}$
\end{algorithmic}
\end{algorithm}

\section{Comparison with Hierarchical Gradient Blending}
\label{sec:sup_hgb}

We perform a comparison with FedHGB~\cite{chen2022towards}, and the results are presented in Table~\ref{tab:fedhgb}.
FedHGB requires labeled data to calculate its blending coefficients. Therefore, we use the ADVANCE dataset under the same settings as in Section~\ref{sec:sup_setting}, with concatenation as the fusion approach.
The resource requirement of FedHGB is similar to that of FedMML.

\begin{table}[tb]
\centering
\caption{Comparison with FedHGB~\cite{chen2022towards}.}
\label{tab:fedhgb}
\begin{tabular}{l|cccc}
\toprule
                     & \textbf{Precision} & \textbf{Recall} & \textbf{F1}    & \textbf{Accuracy} \\ \midrule
\textbf{FedHGB}      & 60.01              & 57.32           & 58.31          & 81.64             \\
\textbf{Prog-FedMML} & \textbf{74.79}     & \textbf{72.91}  & \textbf{73.80} & \textbf{95.76}    \\
\textbf{LW-FedMML}   & 74.19              & 72.17           & 73.11          & 95.17             \\ \bottomrule
\end{tabular}
\end{table}

\section{Illustrative Comparison of Different Approaches}
\label{sec:compare_illu}

In Fig.~\ref{fig:three_comparison}, we visually highlight the key differences between the three approaches: FedMML (end-to-end), Prog-FedMML (progressive), and LW-FedMML (layer-wise).

\section{Discussion}

\label{sec:discussion}

\subsection{Choosing between LW-FedMML and Prog-FedMML}

Here, we discuss the decision-making process for selecting between LW-FedMML and Prog-FedMML, as these approaches can significantly influence resource consumption and performance outcomes. 
The choice between them can be based on the potential client participation rate, which is further influenced by the resource requirements of each method.
While measuring exact resource requirements for each approach can be challenging, it is often feasible to estimate them before initiating the training process.
This estimation can be as simple as running both training processes on a sample device.
If device statistics (e.g., type of device) from the targeted clients (e.g., those located in a certain region) are available, the potential client participation ratio can be estimated based on the resource requirements. 
Device statistics can be less sensitive than actual training data and are often accessible in practice.
This allows for determining the participation gap between \mbox{LW-FedMML} and Prog-FedMML. 
If the gap is large, meaning many clients cannot support Prog-FedMML, LW-FedMML would be the preferable choice. 
Conversely, if the gap is small, indicating most clients can handle \mbox{Prog-FedMML}, it would be the more suitable option.

\subsection{Handling New Modalities}

In some cases, a new modality may be introduced to the system after the training stages have been completed. 
For instance, a sensor with a new modality $c$ could be added to an IoT system that already incorporates modalities $a$ and $b$. 
For conventional end-to-end training (i.e., FedMML), we can integrate a new encoder $F_c$ while freezing the previously trained encoders $F_a$ and $F_b$, avoiding the need to restart training from scratch.
To prevent over-reliance on the pre-trained modalities, regularization techniques such as modality dropout~\cite{neverova2015moddrop} can be applied during the training of $F_c$.
By dropping out pre-existing modalities $a$ and $b$, the model is forced to rely on new modality $c$, facilitating more effective training of $F_c$.
This strategy can also be extended to layer-wise and progressive training settings. In LW-FedMML, during each stage $s$, only the layer $L_c^s$ within the new encoder $F_c^s$ is trained, while the encoders for the existing modalities (i.e., $F_a^s$ and $F_b^s$) are kept frozen. 
Similarly, in Prog-FedMML, the layers $[L_c^1, \dots, L_c^s]$ within $F_c$ can be trained at each stage $s$, while keeping the other modality branches frozen.
In this way, both LW-FedMML and Prog-FedMML can incorporate new modalities without requiring a complete retraining of the entire model.

\ifCLASSOPTIONcaptionsoff
  \newpage
\fi



%



\bibliographystyle{IEEEtran}
\bibliography{reference}

%








\end{document}